%% file: paper.tex
\documentclass[10pt,a4paper,logo]{paper}

\usepackage[all]{hypcap}

\usepackage[authoryear, round]{natbib}

\usepackage{nicefrac}       
\usepackage{multirow}
\usepackage{cleveref}
\usepackage{dsfont}
\usepackage{makecell}
\usepackage{subcaption}
\usepackage{bbm}
\usepackage{wrapfig}
\usepackage{mathtools,xparse}
\usepackage{array}
\usepackage{arydshln}
\usepackage{scalerel}
\usepackage{tablefootnote}
\usepackage{tocloft}
\usepackage{adjustbox}
\usepackage{setspace}
\usepackage[most,skins,theorems]{tcolorbox}

\newtcolorbox{promptbox}{
  colback=gray!5,
  colframe=gray!70,
  boxrule=0.8pt,
  arc=2mm,
  left=10pt,
  right=10pt,
  top=10pt,
  bottom=10pt,
  width=\linewidth,
  fontupper=\ttfamily\footnotesize,
  enhanced,
}

\newboolean{showsection}
\setboolean{showsection}{true}
\makeatletter
\@namedef{ver@everyshi.sty}{}
\makeatother

\definecolor{custom_green}{rgb}{0.0, 0.5, 0.0}
\definecolor{custom_red}{rgb}{1.0, 0.01, 0.24}
\definecolor{custom_blue}{HTML}{C9DAF7}
\definecolor{custom_purple}{HTML}{D9D1E9}
\definecolor{title_blue}{HTML}{204899}
\definecolor{cite_blue}{HTML}{044dc1}
\definecolor{cite_purple}{HTML}{7406a7}

\hypersetup{
    colorlinks = true,
    citecolor = {cite_blue},
    linkcolor = {cite_purple},
    urlcolor = {cite_purple},
}

\input{macro}

\input{math_commands.tex}

\makeatletter
\def\mathcolor#1#{\@mathcolor{#1}}
\def\@mathcolor#1#2#3{%
  \protect\leavevmode
  \begingroup
    \color#1{#2}#3%
  \endgroup
}
\makeatother

\Crefformat{equation}{#2Eq.\;(#1)#3}
\Crefformat{figure}{#2Figure #1#3}
\Crefformat{assumption}{#2Assumption #1#3}
\Crefname{assumption}{Assumption}{Assumptions}

\usepackage{crossreftools}
\makeatletter
\renewcommand\footnoterule{%
  \kern 15\p@
  \hrule \@width 2in \kern 2.6\p@
  \vspace{4pt}
}
\makeatother

\title{Adapting Video Models for Robot Control \\ via Motion-guided Asymmetric Prediction}

\reportnumber{}

\author[1 2]{Taeyoung Kim}
\author[1]{Junyoung Sung}
\author[2]{Heeseung Kwon}
\author[2]{Minseop Kim}
\author[1 2]{Jinwoo Shin}

\affil[1]{KAIST}
\affil[2]{RLWRLD}

\correspondingauthor{
    Correspondence to: \email{jinwoos@kaist.ac.kr}
}

\input{sections/0_abstract}

\begin{document}

\maketitle

\input{sections/1_introduction}
\input{sections/2_method}
\input{sections/3_experiments}
\input{sections/4_related_work}
\input{sections/5_conclusion}

\section*{Acknowledgments}
This work was partly supported by Institute for Information \& Communications Technology Planning \& Evaluation (IITP) grant funded by the Korea government (MSIT)
(RS-2019-II190075, Artificial Intelligence Graduate School Program (KAIST)), and RLWRLD Inc.
We thank the members of Algorithmic General Intelligence Laboratory (AGI-LAB) of KAIST, for helpful support in conducting real-world experiments.

\bibliographystyle{plainnat}
\bibliography{paper}

\clearpage
\appendix{\input{sections/appendix}}

\end{document}

%% file: macro.tex
\definecolor{blanchedalmond}{rgb}{1.0, 0.92, 0.8}
\definecolor{carmine}{rgb}{0.59, 0.0, 0.09}
\definecolor{lightblue}{rgb}{0.22,0.45,0.70}%

\renewcommand{\mathbf}{\boldsymbol}

\makeatletter
\def\Ddots{\mathinner{\mkern1mu\raise\p@
\vbox{\kern7\p@\hbox{.}}\mkern2mu
\raise4\p@\hbox{.}\mkern2mu\raise7\p@\hbox{.}\mkern1mu}}
\makeatother

\numberwithin{equation}{section}

\definecolor{amaranth}{rgb}{0.9, 0.17, 0.31}
\definecolor{antiquebrass}{rgb}{0.8, 0.58, 0.46}
\definecolor{antiquefuchsia}{rgb}{0.57, 0.36, 0.51}
\definecolor{chromeyellow}{rgb}{0.31, 0.47, 0.26}

\newcommand{\1}{\mathds 1}


%% file: math_commands.tex
\usepackage{amsmath,amsfonts,bm}

\def\eqref#1{equation~\ref{#1}}

\def\1{\bm{1}}

\def\rva{{\mathbf{a}}}

\def\rvc{{\mathbf{c}}}

\def\rvo{{\mathbf{o}}}

\def\rvq{{\mathbf{q}}}

\def\rvz{{\mathbf{z}}}

\DeclareMathAlphabet{\mathsfit}{\encodingdefault}{\sfdefault}{m}{sl}
\SetMathAlphabet{\mathsfit}{bold}{\encodingdefault}{\sfdefault}{bx}{n}



%% file: sections/0_abstract.tex
\begin{abstract}
World Action Models (WAMs) leverage pretrained video generative models for robotic control by jointly predicting future visual observations and actions.
The visual prediction objective plays a key role in adapting pretrained video representations to the target robot domain.
Yet standard visual prediction inherited from video pretraining treats all visual regions uniformly, despite their differing relevance to robotic control.
We therefore investigate the spatial allocation of visual supervision in WAMs and show that focusing supervision on dynamic regions can substantially improve policy performance.
Building on this finding, we introduce \textit{Motion-Guided Asymmetric Prediction (MoGAP)}, which augments WAM training with motion-focused visual prediction.
MoGAP allocates noise asymmetrically across regions, training the model to reconstruct more heavily corrupted dynamic regions using less corrupted static regions as context.
MoGAP is trained alongside standard video–action prediction through a training-only auxiliary prediction path within the shared video diffusion backbone. 
It uses motion cues derived directly from training videos, requiring neither an external teacher nor additional modalities.
Across robot manipulation benchmarks and WAM architectures, MoGAP consistently improves policy performance, increasing success rates on the GR-1 tabletop benchmark from 38.8\% to 47.5\% for Fast-WAM (Wan2.2-5B) and from 52.3\% to 62.3\% for Cosmos3-Nano. 
\end{abstract}

%% file: sections/1_introduction.tex
\section{Introduction}
World Action Models (WAMs) have emerged as a promising approach to robotic foundation models, demonstrating that dense dynamics knowledge acquired through video pretraining can be adapted for robot control.
Built on pretrained video diffusion backbones, WAMs jointly predict future visual observations and robot actions, extending video models’ capacity for dense world modeling to action learning~\citep{pai2025mimic, liang2025video, kim2026cosmos, ye2026world}.

The visual prediction objective plays a central role in this adaptation by supervising world representation learning in the target domain.
Removing it has been shown to substantially degrade control performance~\citep{yuan2026fast}, while
jointly optimizing visual and action prediction further shapes these representations for control~\citep{ma2026dit4dit}.
Yet, the objective typically retains the formulation inherited from video pretraining, which was designed for visual generation~\citep{lipman2022flow, liu2022rectified} rather than robot control.

Recent approaches seek to bridge this gap by introducing additional semantic~\citep{qiu2026making,wang2026st} or geometric supervision~\citep{guo2026unified,zhang2026learning,zhao2026rynnworld}. 
However, these approaches primarily rely on external pretrained models or incorporate additional modalities such as depth or 3D information.
We instead ask how visual supervision within the pretrained video backbone’s native latent space can be tailored to better support action learning.

\input{figures/fig1_motivation}

Our motivating study shows that shifting visual supervision toward dynamic regions can consistently improve policy performance across WAMs~(\Cref{subsec:mfm}).
We interpret these gains as evidence that emphasizing dynamic regions can better align visual supervision with the demands of robot control.
Dynamic regions tend to capture robot motion and object interactions, whereas static regions may contain visually complex but less action-relevant details, such as background textures.
These findings motivate assigning distinct roles to dynamic and static regions in visual prediction for action learning.

Building on this insight, we propose \textit{Motion-Guided Asymmetric Prediction (MoGAP)}, a new visual prediction task for WAMs that treats dynamic regions as prediction targets and uses static regions as scene context.
MoGAP establishes this distinction through asymmetric noise allocation~\citep{chen2024diffusion, chefer2026self}, applying higher noise levels to dynamic regions while preserving more information in static regions.
The noise allocation is guided by latent motion scores directly computed from temporal differences in training videos, requiring neither external teacher models nor additional supervision modalities.
MoGAP is jointly optimized with the standard WAM objective through a prediction path within the shared video diffusion backbone. 
This path is used only during training, leaving inference unchanged.

Through extensive experiments, we demonstrate that MoGAP consistently improves policy performance across WAM architectures in both simulated and real-world manipulation benchmarks. 
On GR-1 Tabletop, MoGAP increases the success rate of Fast-WAM (Wan2.2-5B) from 38.8\% to 47.5\% and that of Cosmos3-Nano from 52.3\% to 62.3\%, outperforming competitive VLA and WAM baselines as well as representation alignment methods using external teachers.
On a real-world OpenArm humanoid, we find that MoGAP helps the policy better respond to task-relevant perturbations, such as target object relocation during execution, even without demonstrations of such perturbations.

In summary, our contributions are as follows:
\vspace{-0.5em}
\begin{itemize}[leftmargin=*,itemsep=0mm]
    \item Through an intuitive motivating study, we show that the spatial allocation of visual supervision has a substantial impact on downstream control performance, establishing it as an important design consideration for WAMs.
    
    \item We propose \textit{Motion-Guided Asymmetric Prediction (MoGAP)}, a visual prediction task that trains WAMs to predict more heavily corrupted dynamic regions from cleaner static context, without requiring external supervision models or additional modalities.
    
    \item Through extensive experiments, we demonstrate that MoGAP delivers consistent and substantial gains over strong WAM baselines across diverse tasks, embodiments, and environments.
\end{itemize}

%% file: figures/fig1_motivation.tex
\begin{figure}[t]
    \centering\small
    \includegraphics[width=\linewidth]{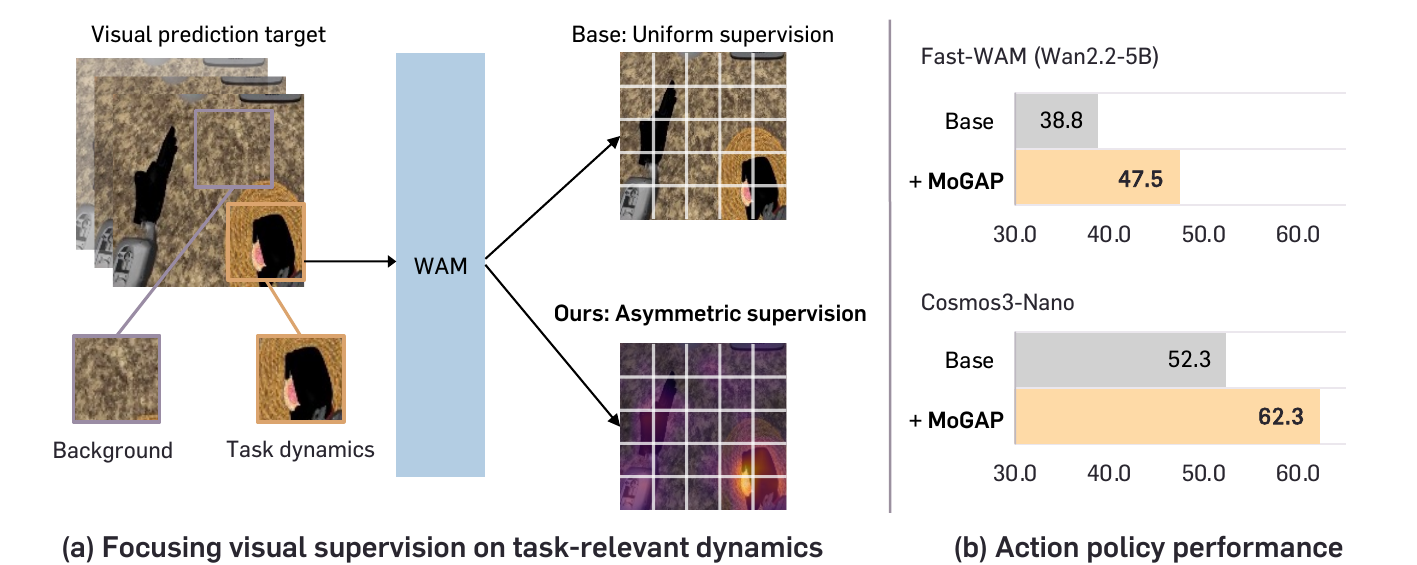}
    \caption{\textbf{Overview.}
    (a) We spatially reallocate visual supervision in World Action Models (WAMs) toward dynamic regions.
    This reallocation emphasizes robot motion and object interactions while downweighting static, task-irrelevant visual details, such as complex tabletop textures.
    (b) Our method, Motion-Guided Asymmetric Prediction (MoGAP), substantially improves policy performance across different WAM setups on GR-1 Tabletop, a challenging humanoid manipulation benchmark.
    }
\label{figure:motivation}
\end{figure}

%% file: sections/2_method.tex
\section{Method}
\label{sec:method}

\input{figures/fig2_method}
We introduce \textit{Motion-Guided Asymmetric Prediction (MoGAP)}, a visual prediction task that promotes control-relevant representation learning in World Action Models (WAMs).
We first describe the standard WAM framework in \Cref{subsec:wam}.
In \Cref{subsec:mfm}, we present a motivating study investigating how the visual prediction objective affects policy learning.
Finally in \Cref{subsec:mdts}, we present our \textit{MoGAP}, with an illustrated overview in \Cref{figure:method}.

\subsection{Preliminaries: World Action Models}
\label{subsec:wam}

At timestep $t$, World Action Models (WAMs) predict future visual observations $\rvo_{t+1:t+H+1}$ and an action chunk $\rva_{t:t+H}$ conditioned on the current observation $\rvo_t$, task instruction $\rvc$, and proprioceptive state $\rvq_t$.
Here, slice notation denotes half-open intervals, so both prediction sequences have horizon $H$.
The model $f_\theta$ learns the conditional joint distribution
\begin{equation}
    p_\theta\!\left(
        \rvo_{t+1:t+H+1}, \rva_{t:t+H}
        \mid \rvo_t, \rvc, \rvq_t
    \right).
\end{equation}
Building on pretrained video diffusion backbones, WAMs perform visual prediction in the latent space of a pretrained video autoencoder.
We denote the encoded future observations by $\rvz_t = \mathcal{E}(\rvo_{t+1:t+H+1})$, where $\mathcal{E}$ is the video encoder; the corresponding decoder maps predicted latents back to the observation space.

Recent approaches~\citep{kim2026cosmos,ye2026world,yuan2026fast,agarwal2026cosmos} train visual and action prediction jointly through flow matching~\citep{lipman2022flow, liu2022rectified}.
At flow-matching timesteps $\tau_v,\tau_a \in [0,1]$, noisy visual latents and actions are constructed as
\begin{align}
    \rvz_t^{\tau_v}
    = (1-\tau_v)\rvz_t
       + \tau_v\boldsymbol{\epsilon}_v, \quad
    \rva_{t}^{\tau_a}
    = (1-\tau_a)\rva_{t:t+H}
       + \tau_a\boldsymbol{\epsilon}_a,
\end{align}
where $\boldsymbol{\epsilon}_v,\boldsymbol{\epsilon}_a \sim \mathcal{N}(\mathbf{0},\mathbf{I})$ are sampled independently.
The model predicts visual and action velocity fields, denoted by $(\hat{\mathbf{u}}_v,\hat{\mathbf{u}}_a)=f_\theta(\rvz_t^{\tau_v},\rva_t^{\tau_a};\rvo_t,\rvc,\rvq_t)$, and is trained with the flow-matching objectives:
\begin{align}
    \mathcal{L}_{\mathrm{video}}
    &=
    \mathbb{E}_{\rvz,\rva}\left[
    \left\|
    \hat{\mathbf{u}}_v
    - (\boldsymbol{\epsilon}_v-\rvz_t)
    \right\|_2^2
    \right],
    \label{eq:video_fm}
    \\
    \mathcal{L}_{\mathrm{action}}
    &=
    \mathbb{E}_{\rvz,\rva}\left[
    \left\|
    \hat{\mathbf{u}}_a
    - (\boldsymbol{\epsilon}_a-\rva_{t:t+H})
    \right\|_2^2
    \right],
    \label{eq:action_fm}
\end{align}
The overall training objective is
\begin{equation}
    \mathcal{L}
    =
    \mathcal{L}_{\mathrm{video}}
    +
    \lambda_{\mathrm{action}}
    \mathcal{L}_{\mathrm{action}},
    \label{eq:wam_loss}
\end{equation}
where $\lambda_{\mathrm{action}} > 0$ is a weighting coefficient that balances the two objectives.

\subsection{Motivating Study}
\label{subsec:mfm}
To investigate how visual supervision affects action learning, we conduct a small experiment fixing the supervision budget, and redistributing it spatially across tokens.
For this, we compute a motion proxy signal directly from temporal differences in the clean video latents $\rvz_t$.
For temporal slice $n$ and spatial latent location $p$, we compute a motion score $m_{n,p}=\|\rvz_{t,n,p}-\rvz_{t,n-1,p}\|_2$.
We normalize the scores by their spatial mean and define spatial reallocation weights as follows:
\begin{equation}
w_{n,p}
=
\frac{\varepsilon+\widetilde m_{n,p}^{\beta}}
{\sum_{q=1}^{P}
\left(\varepsilon+\widetilde m_{n,q}^{\beta}\right)/P},
\qquad
\widetilde m_{n,p}
=
\frac{m_{n,p}}
{\sum_{q=1}^{P}m_{n,q}/P}.
\label{eq:motion_weights}
\end{equation}
where $\widetilde m_{n,p}$ is the normalized score, and $P$ is the number of spatial locations.
Larger motion scores receive higher weights, directing more visual supervision toward regions with greater temporal change.
Here, $\beta>0$ controls the emphasis on the high-change regions, while $\varepsilon>0$ maintains supervision in low-change regions.

We use these weights to define the dynamic-region prioritized visual objective:
\begin{equation}
\mathcal{L}_{\mathrm{video}}'
=
\mathbb{E}_{\rvz,\rva}\left[
\frac{1}{NP}
\sum_{n=1}^{N}\sum_{p=1}^{P}
w_{n,p}\,\mathcal{L}_{\mathrm{video}}^{n,p}
\right],
\label{eq:mfm}
\end{equation}
where $\mathcal{L}_{\mathrm{video}}^{n,p}$ denotes the flow-matching error at temporal slice $n$ and spatial location $p$, and $N$ is the number of future latent slices.

\input{figures/fig3_motivating_study}
For this study, we replace $\mathcal{L}_{\mathrm{video}}$ in \Cref{eq:wam_loss} with $\mathcal{L}_{\mathrm{video}}'$ to redistribute visual supervision toward dynamic regions, leaving the action prediction objective unchanged.
As shown in \Cref{fig:motivatingstudy}, this reallocation improves the action policy performance on GR-1 tabletop, a simulated humanoid manipulation benchmark.
We interpret this finding as evidence that visual regions need not contribute equally to action learning: dynamic regions tend to capture robot motion and object interactions, whereas static regions may contain visually complex but less control-relevant details, such as tabletop textures.
This suggests that visual prediction can better support control by assigning different learning roles to dynamic and static regions.

\subsection{Motion-Guided Asymmetric Prediction}
\label{subsec:mdts}
Building on this insight, we introduce \textit{Motion-Guided Asymmetric Prediction (MoGAP)}, a new visual prediction task for WAMs that uses less corrupted static regions as context for predicting more heavily corrupted dynamic regions.
While our motivating experiment prioritizes dynamic regions by reweighting prediction errors, MoGAP extends this principle to the prediction task itself by controlling how much visual information is available in dynamic and static regions.

\vspace{0.01in}
\textbf{Designing the asymmetric prediction task.}~
To construct this asymmetric prediction task, we exploit the interpretation of diffusion noising as partial masking~\citep{chen2024diffusion}, where higher noise levels reveal less information about the underlying clean context.
Given the base visual timestep $\tau_v$, we independently draw $\xi\sim\mathcal{T}_v$, where $\mathcal{T}_v$ is the video timestep sampling distribution, and rescale it to the interval $[\tau_v,1]$:
\begin{equation}
\tau_v^+
=
\tau_v+(1-\tau_v)\xi.
\label{eq:mdts_higher}
\end{equation}
This preserves the base noise level while increasing corruption in dynamic regions, with the noise gap scaled by the remaining interval $1-\tau_v$.
We construct a binary motion mask $M_{n,p}$ by thresholding latent motion scores.
The mask is mapped to the model's spatial token grid to construct a dual-timestep assignment $\boldsymbol{\tau}_M$:
\begin{equation}
M_{n,p}=\mathbb{I}\!\left[m_{n,p}>\bar m_n\right],
\qquad
\tau_{n,p}
=
\begin{cases}
\tau_v^+, & M_{n,p}=1,\\
\tau_v, & M_{n,p}=0,
\end{cases}
\label{eq:mdts_assignment}
\end{equation}
where $\bar m_n$ is the spatial mean.
Using Gaussian noise $\boldsymbol{\epsilon}_v'\sim\mathcal{N}(\mathbf{0},\mathbf{I})$ sampled independently of $\boldsymbol{\epsilon}_v$, we construct the asymmetrically noised visual latents as
\begin{equation}
\rvz_t^{\boldsymbol{\tau}_M}
=
\operatorname{diag}(\mathbf{1}-\boldsymbol{\tau}_M)\rvz_t
+
\operatorname{diag}(\boldsymbol{\tau}_M)\boldsymbol{\epsilon}_v'.
\label{eq:mdts_input}
\end{equation}
The corresponding flow-matching target is $\boldsymbol{\epsilon}_v'-\rvz_t$.
Dynamic regions are thus predicted under greater corruption, with less corrupted static regions providing scene context.

For these noisy inputs, we apply the video flow-matching loss in \Cref{eq:video_fm} only to moving tokens with $M_{n,p}=1$.
Within each latent slice, we normalize by the number of selected locations so that the moving-region fraction does not change the total loss weight.
Let $\mathcal{S}_n=\{p\mid M_{n,p}=1\}$, with all spatial locations selected when the mask is empty.
The resulting objective is: 
\begin{equation}
\mathcal{L}_{\mathrm{MoGAP}}
=
\mathbb{E}_{\rvz,\rva}\left[
\frac{1}{N}\sum_{n=1}^{N}
\frac{1}{|\mathcal{S}_n|}
\sum_{p\in\mathcal{S}_n}
\left\|
\hat{\mathbf{u}}_{v,n,p}^{\boldsymbol{\tau}_M}
-
(\boldsymbol{\epsilon}'_{v,n,p}-\rvz_{t,n,p})
\right\|_2^2
\right],
\label{eq:mdts_loss}
\end{equation}
where $\hat{\mathbf{u}}_{v,n,p}^{\boldsymbol{\tau}_M}$ denotes the visual velocity prediction under the asymmetric timestep assignment.

\input{figures/fig4_attention}
\vspace{0.01in}
\textbf{Incorporating MoGAP into WAMs.}~
As MoGAP provides complementary motion-guided supervision to standard video-action prediction, we incorporate it as an auxiliary training path that augments the original WAM objective.
We process the dual-timestep-noised visual latents $\rvz_t^{\boldsymbol{\tau}_M}$ alongside the original WAM inputs in the shared video diffusion backbone, without introducing additional parameters.
To keep the original WAM prediction path independent of the auxiliary tokens, we apply an asymmetric attention mask between the two paths. 
As shown in ~\Cref{fig:mdtsattn}, the original WAM tokens, including noisy future visual and action tokens, cannot attend to the MoGAP tokens.
MoGAP tokens can attend to one another, the current observation, and noisy action representations, but not the original path’s noisy future visual tokens.
The task instruction $\rvc$ conditions both paths through the existing conditioning mechanism.

\vspace{0.01in}
We jointly optimize the standard WAM objectives with the \mbox{MoGAP} objective:
\vspace{0.01in}
\begin{equation}
\mathcal{L}
=
\mathcal{L}_{\mathrm{video}}
+
\lambda_{\mathrm{action}}\mathcal{L}_{\mathrm{action}}
+
\lambda_{\mathrm{MoGAP}}\mathcal{L}_{\mathrm{MoGAP}},
\label{eq:total_loss}
\end{equation}
where $\lambda_{\mathrm{MoGAP}}>0$ controls the contribution of motion-guided supervision.
At inference time, the MoGAP tokens are omitted, leaving the original WAM inference procedure unchanged.

%% file: figures/fig2_method.tex
\begin{figure}[t]
    \centering\small
    \includegraphics[width=\textwidth]{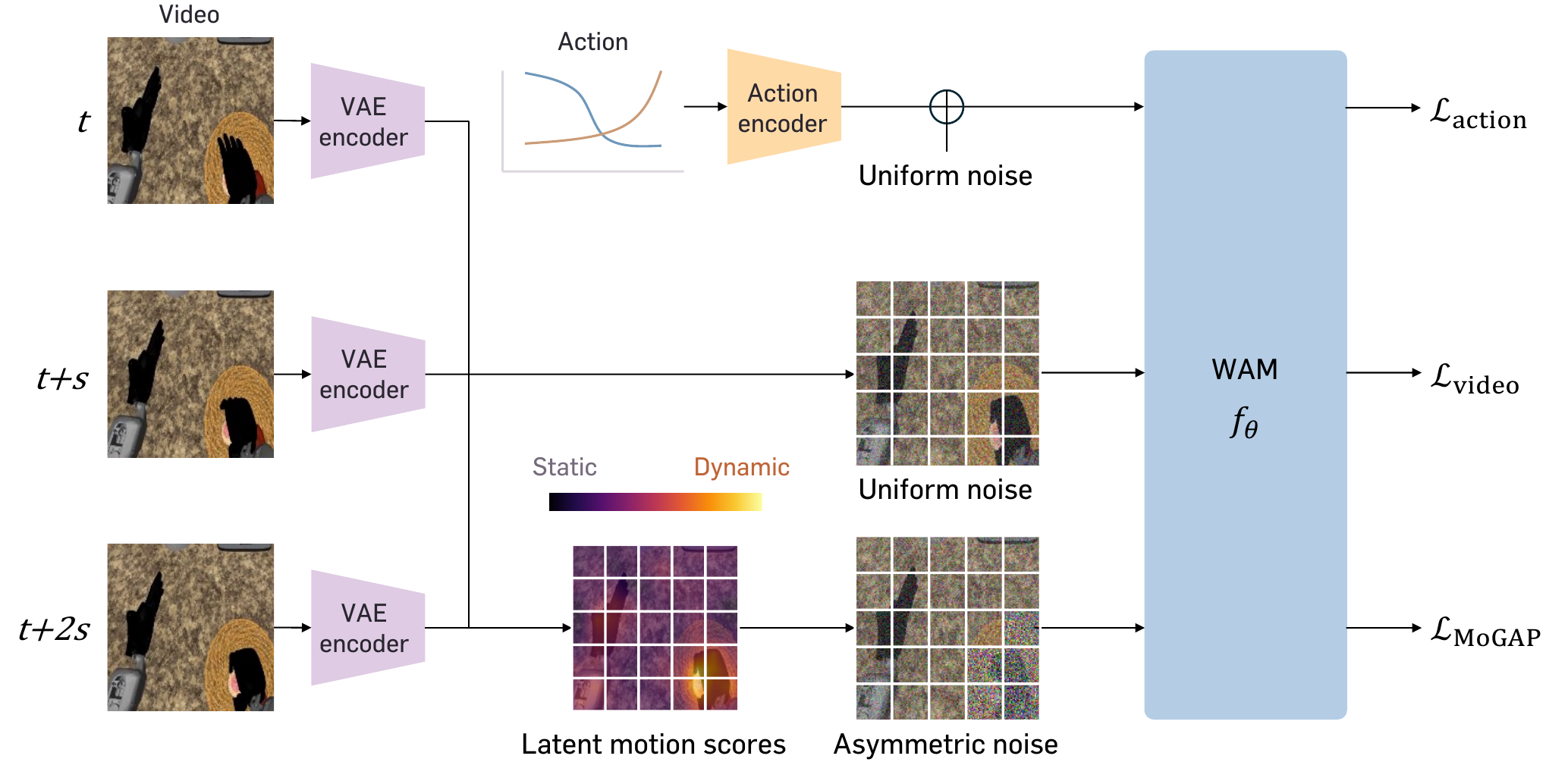}
    \caption{\textbf{Motion-Guided Asymmetric Prediction (MoGAP).}
    MoGAP is a visual prediction task that promotes control-relevant representation learning in World Action Models.
    MoGAP introduces regional asymmetry in noise levels, guided by temporal differences between video latents: dynamic regions receive higher noise levels, while less corrupted static regions serve as context for prediction.
    }
    \label{figure:method}
\end{figure}

%% file: figures/fig3_motivating_study.tex
\begin{wrapfigure}[13]{R}{0.40\textwidth}
\centering\small
\vspace{-2.0em}
\resizebox{0.9\linewidth}{!}{
\includegraphics[width=\linewidth]{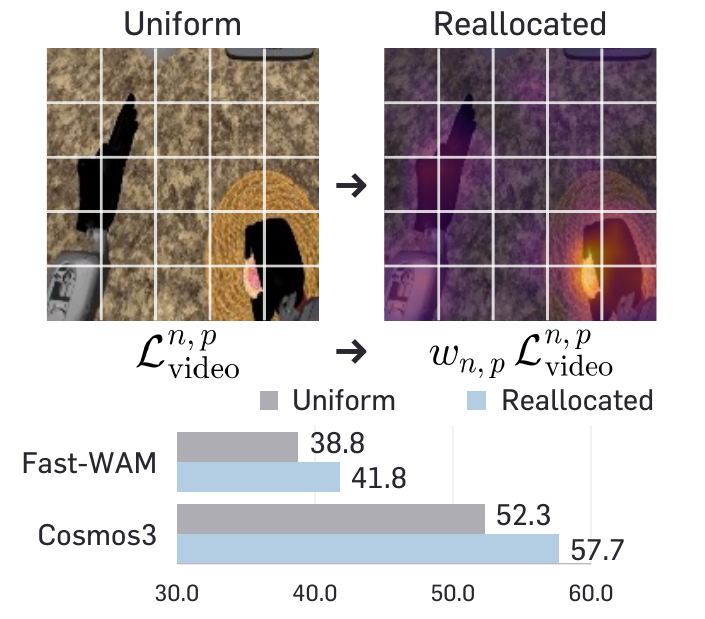}
}
\vspace{-1.0em}
\captionof{figure}{
    \textbf{Motivating study.} Redistributing supervision toward dynamic regions.
}
\label{fig:motivatingstudy}
\end{wrapfigure}

%% file: figures/fig4_attention.tex
\begin{wrapfigure}[16]{R}{0.4\textwidth}
\centering\small
\vspace{-2.0em}
\resizebox{0.36\textwidth}{!}{
\includegraphics[width=0.9\linewidth]{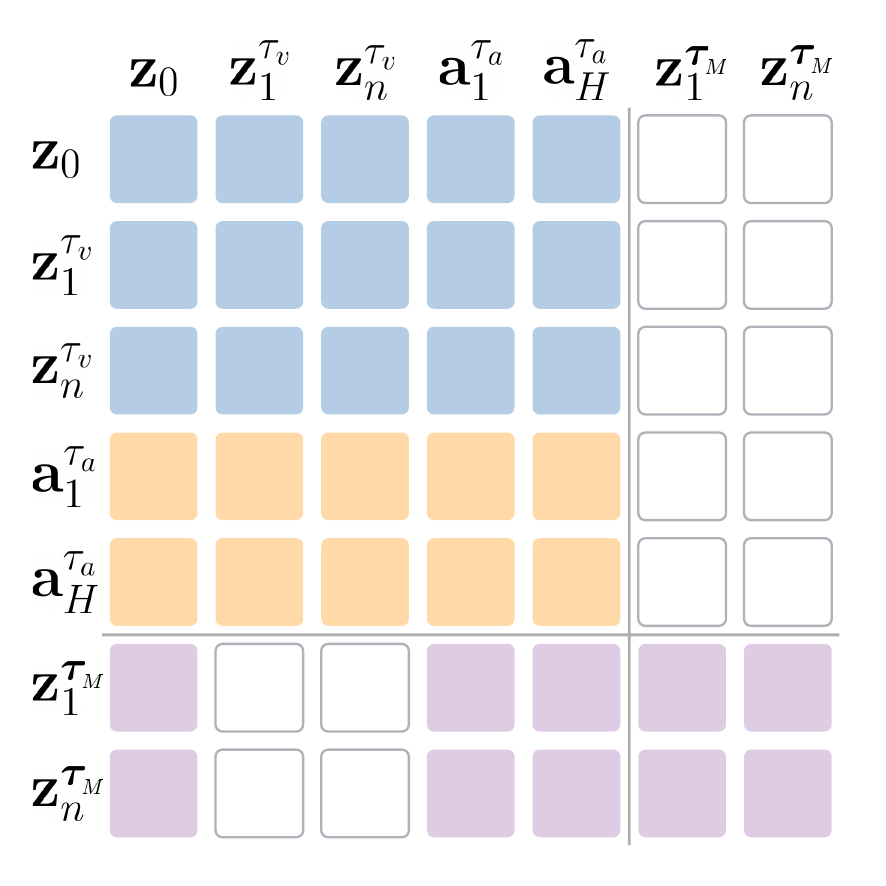}
}
\vspace{-0.5em}
\captionof{figure}{
    \textbf{Attention mask of \mbox{MoGAP}.}
    The MoGAP tokens $\rvz_t^{\boldsymbol{\tau}_M}$  are omitted at inference time.
}
\label{fig:mdtsattn}
\end{wrapfigure}

%% file: sections/3_experiments.tex
\section{Experiments}

In this section, we evaluate MoGAP across diverse manipulation settings, robot embodiments, and WAM architectures.
We first describe the benchmarks and implementation details in \Cref{subsec:setup}.
In \Cref{subsec:results}, we study its impact on frontier WAM baselines, across multitask simulated manipulation benchmarks.
We then assess MoGAP's performance in real-world physical deployment, on an OpenArm Humanoid equipped with Inspire hands, in \Cref{subsec:realresults}.
Finally, in \Cref{subsec:analysis}, we present informative analyses to better understand the effects of MoGAP.

\subsection{Setup}
\label{subsec:setup}

\vspace{0.01in}
\textbf{Benchmarks.}~ We evaluate MoGAP on two multitask simulation benchmarks and a real-world manipulation setup, covering different embodiments, observation configurations, and environmental conditions.
\vspace{-0.5em}
\begin{itemize}[leftmargin=*,itemsep=0mm]
    \item \textbf{RoboCasa GR-1 Tabletop}~\citep{bjorck2025gr00t} consists of 24 tabletop hand manipulation tasks using the GR-1 humanoid. 
    We use this benchmark for our main study, including comparisons with existing methods and ablation studies.

    \item \textbf{RoboCasa-Kitchen}~\citep{nasiriany2024robocasa} comprises 24 manipulation tasks using a single-arm robot with a gripper in kitchen environments. We use this benchmark to evaluate MoGAP in multi-view observation settings.
    
    \item \textbf{Real-world evaluation} uses an OpenArm humanoid with two Inspire RH56F1 hands. We evaluate both in-domain tasks and diverse out-of-domain conditions to assess the abilities of MoGAP in physical deployment.
    
\end{itemize}

\textbf{Implementation details.}~
We adopt Cosmos3-Nano~\citep{agarwal2026cosmos} as our primary WAM, generating 16 actions and 8 future video frames with a temporal stride of two in a single forward process.
We train for 16K gradient steps for the multi-task simulation benchmarks and 8K steps for real-world experiments, respectively, with a global batch size of 256.
We follow the original work for the training details, and set $\lambda_{\mathrm{MoGAP}} = 1.0$.
For the simulated benchmarks, we evaluate each method over 50 trials per task, for a total of 1,200 rollouts.
Further details are described in Appendix~\ref{appendix:implementationdetails}.

\input{tables/tab1_gr1tabletop}
\input{tables/tab2_comparison}

\subsection{Simulated benchmark results}
\label{subsec:results}

\vspace{0.01in}
\textbf{Comparison with state-of-the-art methods.}~
\Cref{tab:gr1_main} summarizes the performance on RoboCasa GR-1 Tabletop.
Cosmos3 already provides a strong baseline, outperforming the other WAM-based methods and remaining competitive with leading VLA-based methods.
Despite this strong starting point, MoGAP improves the average success rate from 52.3\% to 62.3\%, a gain of 10.0 percentage points, surpassing all VLA and WAM-based methods in the comparison.
The gains are particularly pronounced on pick-and-place tasks, which require precise manipulation of a wider variety of objects within each task, with success rates increasing from 48.3\% to 60.8\% (+12.5 percentage points).
We attribute this improvement to MoGAP's emphasis on fine-grained motion that produces only subtle visual changes but is important for manipulation, especially for precise positioning during grasping and placing.
We further visualize this effect in our qualitative analysis (see~\Cref{fig:analysis}).

\vspace{0.01in}
\textbf{Comparison to teacher-based supervision.}~
We compare MoGAP with teacher-based representation learning methods that leverage visual foundation models~\citep{oquab2023dinov2,wang2023videomae, mur2026v} or 3D foundation models~\citep{wang2025vggt} as teachers, using the same Cosmos3-Nano baseline model.
As shown in~\Cref{tab:comparison}, all the other methods improve over the baseline, yet MoGAP outperforms them without an external teacher.
While external representations provide useful supervision, MoGAP shows that motion-guided regional noise allocation offers an effective alternative for adapting the pretrained video backbone's native representations to robot control.

\input{tables/tab3_ablations}

\textbf{Component analysis.}
\Cref{tab:component_analysis} studies individual components of MoGAP.
As shown in the first two rows, replacing the main visual supervision with our method improves the success rate from 52.3\% to 54.4\%, while introducing it as an auxiliary branch yields a substantially larger gain.
This suggests that the standard full-noise range supervision and our asymmetric supervision provide complementary learning signals.
The third and fourth rows further show that removing the noise allocation asymmetry or reversing the asymmetry direction significantly reduces the success rate to 55.3\% and 55.2\%, respectively.
These results demonstrate that asymmetric noise allocation toward predicting dynamic regions from static context is a crucial design choice for downstream performance.

\vspace{0.01in}
\textbf{Generalization across WAM architectures.}~
We further assess MoGAP on Fast-WAM~\citep{yuan2026fast}, which combines a Wan2.2-5B~\citep{wan2025wan} video diffusion backbone with a separately parameterized action expert and omits future visual prediction at inference.
MoGAP improves the success rate from 38.8\% to 47.5\% (\Cref{tab:fastwam}).
Together with the improvements on Cosmos3, these results demonstrate MoGAP's effectiveness across different WAM architectures and video backbones.

\input{figures/fig5_robocasakitchen}

\vspace{0.01in}
\textbf{Adaptation to multi-view configuration.}~
We also evaluate MoGAP on RoboCasa-Kitchen, a single-arm manipulation benchmark with three camera views.
We apply motion-guided noise assignment to the two external camera views while assigning all wrist-camera tokens to the high-noise group, as camera motion can dominate the latent motion scores in the wrist view.
With this adaptation, MoGAP consistently outperforms the Cosmos3 baseline while retaining the benefits of longer execution horizons (\Cref{fig:robocasa}).
These results demonstrate that MoGAP extends to multi-view settings that include a moving wrist camera.

\subsection{Real-world experiment results}
\label{subsec:realresults}

\vspace{0.01in}
\textbf{Real-world experimental setup.}~
We conduct experiments on an OpenArm humanoid with two 7-DoF arms and two five-finger, 6-DoF hands and an egocentric stereo camera.
We evaluate three task groups targeting different aspects of policy performance: \textit{In-Domain} tasks for general manipulation, \textit{OOD Objects \& Environments} tasks for robustness to unseen objects and environments, and \textit{OOD Object Relocation} tasks for evaluating the policy's ability to focus on task-relevant regions (\Cref{fig:realworld_task}).
For low-latency deployment, we use a single denoising step for both visual and action prediction without classifier-free guidance~\citep{ho2022classifier}.
We precompute and cache text embeddings, removing the text encoder from the inference path, and omit the VAE decoder.
Together, these choices enable Cosmos3-Nano, a 16B-parameter model, to run with an inference latency of 48ms on a single RTX 5090 GPU with 32GB of VRAM.
For each method, we train a single policy on the same multitask dataset collected in our real-world setup.
Further details are provided in Appendix~\ref{appendix:realdetails}.

\vspace{0.01in}
\textbf{Real-world evaluation results.}~
\Cref{fig:realworld_result} presents the results on the real-robot manipulation tasks. 
MoGAP consistently improves over the baseline policy across all three groups, demonstrating gains in both general manipulation and robustness to distribution shifts.
Notably, we find that MoGAP yields particularly large gains on the object relocation tasks.
In these trials, the baseline policy often continues its original trajectory and attempts to grasp at the object's previous location, whereas the MoGAP-trained policy more successfully adjusts its trajectory toward the relocated object.
This adaptation emerges despite the absence of object relocation demonstrations during training.
We attribute this behavior in part to MoGAP's emphasis on dynamic regions, which encourages the policy to respond more effectively to task-relevant visual changes in the environment.

\input{figures/fig6_openarm}
\input{figures/fig7_openarm_relocation}

\subsection{Further Analysis}
\label{subsec:analysis}

\vspace{0.01in}
\textbf{Motion score analysis.}~
We analyze the motion score derived from temporal differences between video latents.
In \Cref{fig:analysis}, the scores highlight subtle arm and hand movements (rows 1 and 2) and remain focused on the robot and objects even under camera shifts caused by waist motion (row 3).
We quantify this localization using region-wise AUROC, which evaluates how well the motion scores rank locations within each ground-truth region relative to the rest of the scene (\Cref{tab:motionscore}).
The results show that the signal clearly assigns higher scores to the moving arm (including the hand) and the moving target object, while assigning relatively lower scores to background regions (i.e., the tabletop, surrounding environment, and distractor objects).
\input{figures/fig8_analysis}
\input{tables/tab4_analysis}

\vspace{0.01in}
\textbf{Trained policy analysis.}~
Training with the latent motion score supervision encourages WAMs to attend more closely to dynamic regions during action generation and improves their prediction of these regions.
In~\Cref{fig:analysis}, we visualize the action tokens' attention, and observe that it is concentrated more strongly on the moving hand and arm. 
In the generated examples, the baseline generation often appears blurred around the moving arm, while ours preserves its structure relatively more clearly (rows 2 and 3). 
Quantitative results in \Cref{tab:policyanalysis} support this attention shift, showing greater attention to the moving arm and hand and the manipulated object, and reduced attention to the background. 
Further details about the analysis metrics are provided in Appendix~\ref{appendix:analysisdetails}.

\vspace{0.01in}
\textbf{Limitations.}~
MoGAP introduces additional tokens during training, increasing wall-clock training time by a factor of 1.26 in exchange for improved policy performance.
To assess this trade-off, we provide computational details and a compute-matched comparison in Appendix~\ref{appendix:computationdetails}.

%% file: tables/tab1_gr1tabletop.tex
\begin{table}[t]
\centering\small
\caption{
    \textbf{Comparison with state-of-the-art VLAs \& WAMs.} 
    Results include the success rate (\%) of representative VLAs and WAMs on GR-1 tabletop. Results of Fast-WAM and Cosmos3 are reproduced while others are from the original work. 
    Best results are highlighted in \textbf{bold}.
}
\label{tab:gr1_main}
\begin{tabular}{l ccc}
    \toprule
    Method & PnP & Articulated
    & \makecell{24 Tasks\\Average} \\
    \midrule
    \multicolumn{4}{l}{\textit{VLA-based methods}} \\
    \quad GR00T-N1.6~\citep{bjorck2025gr00t}      & 55.4 & 24.2 & 47.6 \\
    \quad StarVLA-$\alpha$~\citep{ye2026starvla} & 58.4 & 54.0 & 57.3 \\
    \quad RLDX-1~\citep{kim2026rldx}           & 59.0 & 57.7 & 58.7 \\
    
    \midrule
    \multicolumn{4}{l}{\textit{WAM-based methods}} \\
    \quad Fast-WAM~\citep{yuan2026fast} & 33.8 & 53.7 & 38.8 \\
    \quad DiT4DiT~\citep{ma2026dit4dit} & 50.9 & 50.3 & 50.8 \\
    \quad Cosmos3~\citep{agarwal2026cosmos} & 48.3 & 64.3 & 52.3 \\
    
    \rowcolor{gray!10}
    \quad \textbf{MoGAP (Ours)} & \textbf{60.8} & \textbf{66.7} & \textbf{62.3} \\
\bottomrule
\end{tabular}
\end{table}

%% file: tables/tab2_comparison.tex
\begin{table}[t!]
\centering\small
\caption{
\textbf{Comparison to teacher-based supervision.}
Results report the average success rate (\%) on GR-1 tabletop, with each method applied to the same baseline WAM, Cosmos3-Nano~\citep{agarwal2026cosmos}.
Best results are highlighted in \textbf{bold}.
}
\label{tab:comparison}
\begin{tabular}{l l c}
\toprule
Method & Teacher &
\makecell{24 Tasks\\Average} \\
\midrule
Cosmos3 & -- & 52.3 \\
\quad w/ Geometry Forcing~\citep{wu2026geometry}
& VGGT & 53.9 \\
\quad w/ AGRA~\citep{qiu2026making} & DINOv2 & 55.4 \\
\quad w/ TSA~\citep{kim2026tempered}
& VideoMAEv2 & 56.8 \\
\quad w/ VideoREPA~\citep{zhang2026videorepa}
& VideoMAEv2 & 57.8 \\
\quad w/ VideoREPA~\citep{zhang2026videorepa}
& V-JEPA2.1-L & 59.2 \\
\rowcolor{gray!10}
\textbf{\quad w/ MoGAP (Ours)} & -- & \textbf{62.3} \\
\bottomrule
\end{tabular}
\end{table}

%% file: tables/tab3_ablations.tex
\begin{table*}[t!]
\centering
\caption{%
\textbf{Ablation studies on MoGAP.}
We study (a) the effectiveness of its key components through ablations and (b) the generalization of MoGAP across WAM architectures.
Results report average success rates (\%) on GR-1 Tabletop.
Best results within each comparison are highlighted in \textbf{bold}.
}
\label{tab:mogap_results}
\vspace{0.2em}
~~~
\begin{subtable}[t]{0.40\textwidth}
\vspace{0pt}
\centering\small
\begin{tabular}{l c}
\toprule
Method
& \makecell{24 Tasks\\Average} \\
\midrule
\rowcolor{gray!10}
\textbf{MoGAP} & \textbf{62.3} \\
\midrule
w/o auxiliary path & 54.4 \\
w/o noise asymmetry & 55.3 \\
reversed direction & 55.2 \\
\bottomrule
\end{tabular}
\vspace{0.5em}
\caption{Component analysis.}
\label{tab:component_analysis}
\end{subtable}
\begin{subtable}[t]{0.55\textwidth}
\vspace{0pt}
\centering\small
\begin{tabular}{l cc}
\toprule
Method &
Backbone
& \makecell{24 Tasks\\Average} \\
\midrule
Fast-WAM & Wan2.2-5B & 38.8 \\
\rowcolor{gray!10}
\textbf{+ MoGAP} & Wan2.2-5B & \textbf{47.5} \\
\midrule
Cosmos3 & Cosmos3-Nano & 52.3 \\
\rowcolor{gray!10}
\textbf{+ MoGAP} & Cosmos3-Nano & \textbf{62.3} \\
\bottomrule
\end{tabular}%
\vspace{0.5em}
\caption{Generalization across WAMs.}
\label{tab:fastwam}
\end{subtable}%

\vspace{-1.0em}
\end{table*}

%% file: figures/fig5_robocasakitchen.tex
\begin{wrapfigure}[10]{R}{0.38\textwidth}
\centering\small
\vspace{-1.5em}
\resizebox{\linewidth}{!}{
\includegraphics[width=\linewidth]{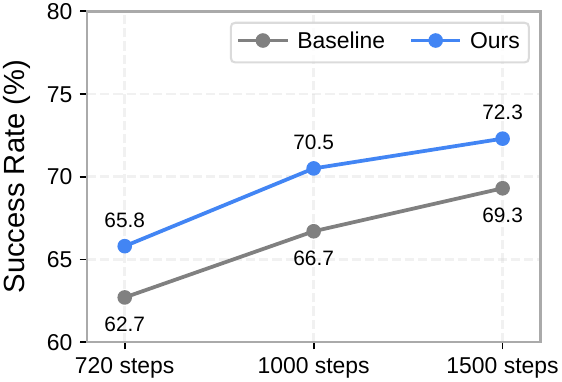}
}
\captionof{figure}{
    \textbf{RoboCasa-Kitchen success rate (\%)} across execution timesteps.
}
\label{fig:robocasa}
\end{wrapfigure}

%% file: figures/fig6_openarm.tex
\begin{figure}[t]
\centering\small
\includegraphics[width=0.95\linewidth]{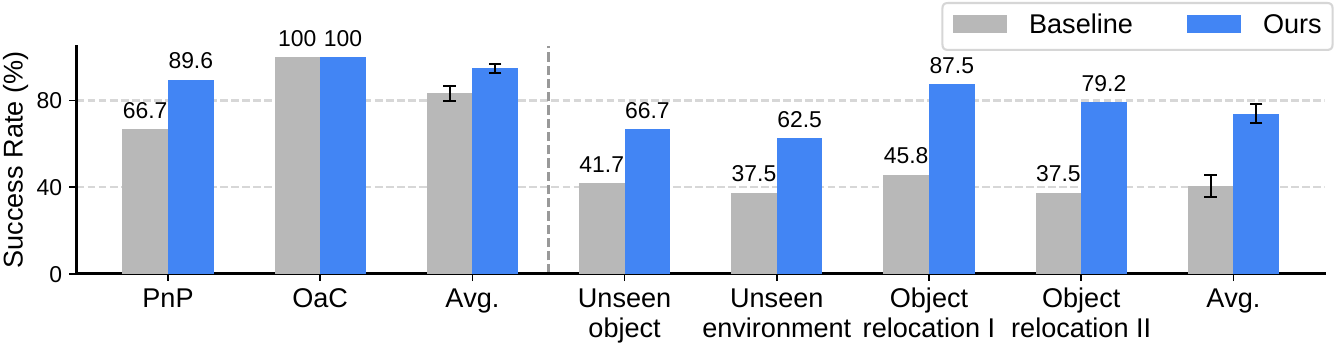}
\captionof{figure}{
\textbf{Real-world experiment success rate (\%).}
Results on the OpenArm humanoid real-world benchmark.
We report success rates with standard errors for each method across task suites.
}
\vspace{-1.0em}
\label{fig:realworld_result}
\end{figure}

%% file: figures/fig7_openarm_relocation.tex
\begin{figure}[t]
\centering\small
\includegraphics[width=.95\linewidth]{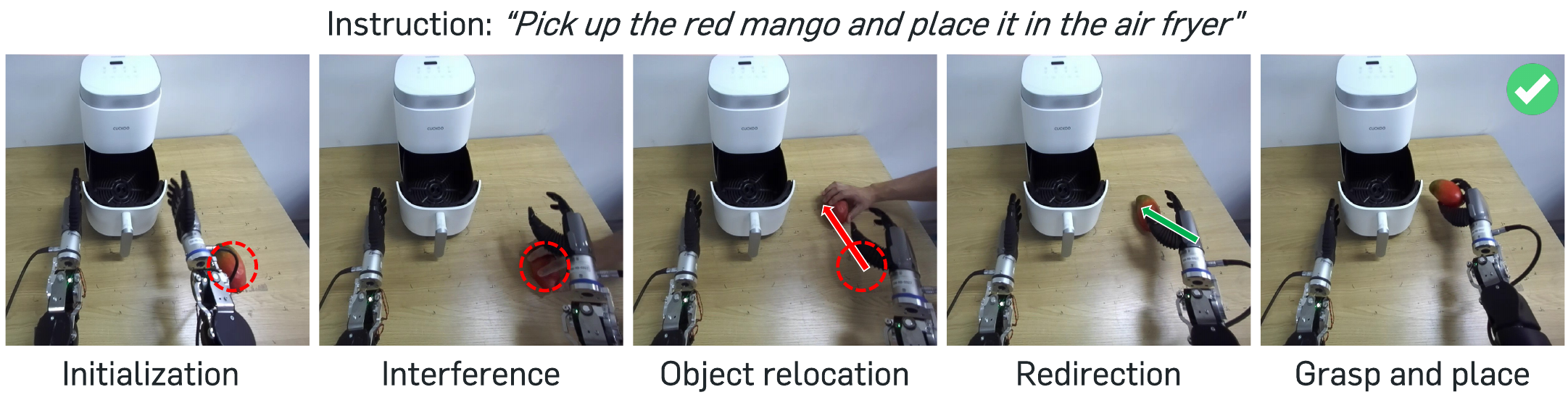}
\captionof{figure}{
\textbf{Object relocation task example.}
The target object's location is moved during execution. In set II, the object is relocated farther than in set I. MoGAP successfully adapts its trajectory.
}
\vspace{-1.0em}
\label{fig:realworld_task}
\end{figure}

%% file: figures/fig8_analysis.tex
\begin{figure}[t]
    \centering\small
    \includegraphics[width=\linewidth]{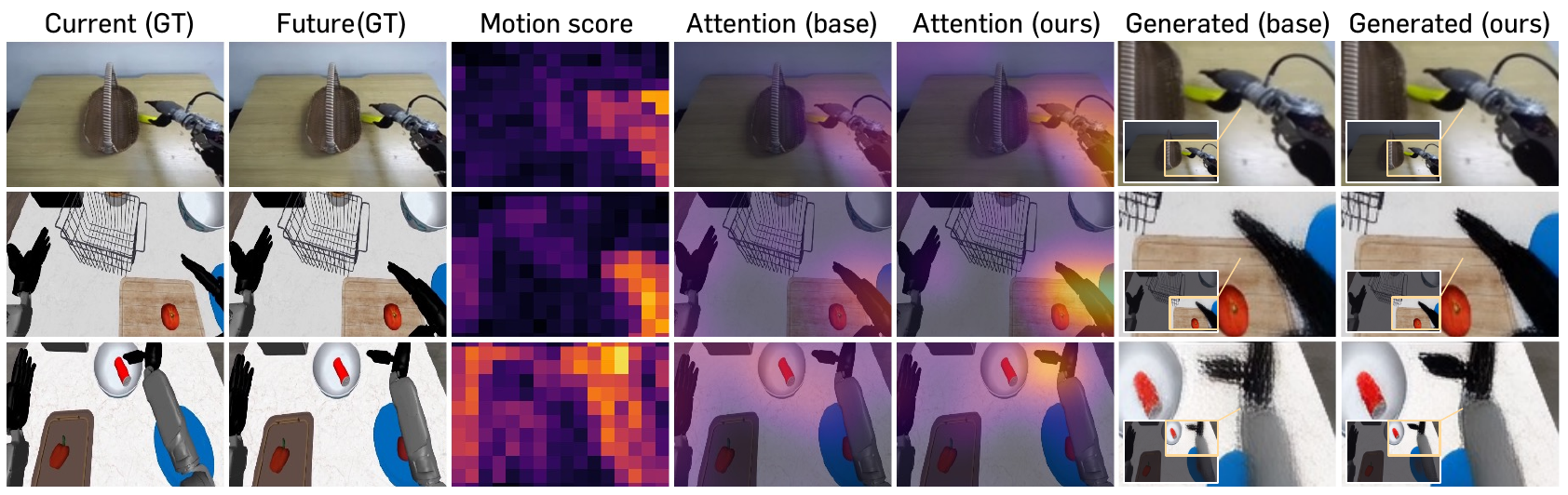}
\captionof{figure}{
\textbf{Qualitative analysis.}
The columns show, from left to right, current and future ground-truth observations, latent motion scores, action-to-vision attention maps for the baseline and MoGAP, and their generated future observations.
}
\label{fig:analysis}
\end{figure}

%% file: tables/tab4_analysis.tex
\begin{table*}[t]
\centering
\caption{\textbf{Quantitative analysis.}
We conduct additional analyses of the latent motion
signal, and the trained policy models on GR-1 Tabletop.}
\label{tab:mogap_analysis}
\vspace{0.2em}

\begin{minipage}[t]{0.38\textwidth}
\vspace{0pt}
\centering\small
\phantomsubcaption\label{tab:motionscore}

\begin{tabular}{l c}
\toprule
\makecell[l]{Evaluation\\Region}
& AUROC \\
\midrule
Moving arm \& hand & 0.887 \\
Manipulated object & 0.834 \\
Background & 0.430 \\
\bottomrule
\end{tabular}%

\vspace{0.6em}
\caption*{\small (a) Motion signal validity.}
\end{minipage}%
~~
\begin{minipage}[t]{0.48\textwidth}
\vspace{0pt}
\centering\small
\phantomsubcaption\label{tab:policyanalysis}

\begin{tabular}{lc}
\toprule
\makecell[l]{Evaluation\\Region}
& \makecell{Attention Ratio\\Base $\rightarrow$ Ours} \\
\midrule
Moving arm \& hand & $1.80 \rightarrow 2.07$ \\
Manipulated object & $2.00 \rightarrow 2.29$ \\
Background & $0.86 \rightarrow 0.82$ \\
\bottomrule
\end{tabular}

\vspace{0.6em}
\caption*{\small (b) Trained policy analysis.}
\end{minipage}
\vspace{-1.0em}
\end{table*}

%% file: sections/4_related_work.tex
\section{Related Work}

\vspace{0.01in}
\textbf{World Action Models (WAMs).}~
WAMs adapt rich spatiotemporal priors from large-scale video data to robot control by learning to predict future visual observations and control actions together~\citep{pai2025mimic, liang2025video, kim2026cosmos, li2026causal, ye2026world}.
However, visual reconstruction alone may not fully align with the needs of robot control~\citep{zhou2024dino,jha2026reconstruction}.
Recent studies incorporate supervision from semantic encoders~\citep{qiu2026making,wang2026st}, or from depth and 3D  geometry~\citep{guo2026unified,zhang2026learning,zhao2026rynnworld, yang20264d}.
Such approaches augment visual supervision through pretrained representation models~\citep{oquab2023dinov2, wang2025vggt}, or additional prediction targets derived from geometry and motion estimators~\citep{lin2025depth, morimitsu2025dpflow, liu2026trace}.
Our work explores a complementary direction by introducing an asymmetric prediction task for action learning within the pretrained video diffusion model's native latent space, without external teachers or additional modalities.

\vspace{0.01in}
\textbf{Representation learning for video generative models.}~
Recent work explores representation learning within video generative models to improve prediction quality.
One line of work aligns diffusion and flow model representations with external pretrained encoders~\citep{yu2025representation}, encouraging physically plausible predictions~\citep{zhang2026videorepa,bhowmik2026moalign} or incorporating 3D geometric structure~\citep{wu2026geometry}.
Another line improves representations through self-supervision without external teachers~\citep{chefer2026self,pan2026temporal}.
Our approach builds on these advances, drawing in particular on token-wise noise assignment~\citep{chen2024diffusion,chefer2026self} to introduce motion-guided asymmetry tailored to action modeling for robot control.

%% file: sections/5_conclusion.tex
\section{Conclusion}
In this work, we present \textit{Motion-Guided Asymmetric Prediction (MoGAP)}, a visual prediction task that adapts pretrained video diffusion models for robot control through motion-guided regional noise allocation.
Our experiments demonstrate consistent improvements over strong WAM baselines in simulated and real-world manipulation settings, including improved adaptation to target object relocation tasks during execution.
These findings highlight the importance of tailoring visual supervision for robot policy learning, particularly by emphasizing regions that capture task-relevant dynamics.
We hope this work encourages further exploration of control-oriented visual prediction  to translate pretrained video models' world knowledge into more capable robotic policies.

%% file: sections/appendix.tex
\newpage
\section{Computational Details}
\label{appendix:computationdetails}
Experiments are conducted on 4 $\times$ NVIDIA H200 Tensor core GPUs. 
In this setup, training Cosmos3 for our GR-1 Tabletop experiments takes 14.7 hours for 16K optimization steps with a global batch size of 256.
MoGAP introduces additional tokens during training, increasing training time by a factor of 1.26 to 18.5 hours.
Caching the VAE latents reduces the training times to 10.9 and 15.0 hours, respectively.
In the main experiments, Cosmos3 and MoGAP are both trained for 16K steps with the same batch size and therefore see the same number of samples. 
For a compute-matched comparison we additionally train Cosmos3 for 22K steps (6K more steps) and report its average success rate in~\Cref{tab:computecomparison}.

\begin{table}[h]
    \centering
    \begin{tabular}{l c}
    \toprule
         Method & \makecell{24 Tasks\\Average} \\
         \midrule
         Cosmos3 (22K) & 55.8\\
         MoGAP  ~(16K) & 62.3 \\
     \bottomrule
    \end{tabular}
    \caption{Compute-matched comparison}
    \label{tab:computecomparison}
\end{table}

\section{Simulation Experiments Details}
\label{appendix:simulationdetails}

\subsection{Dataset}
For RoboCasa GR-1 tabletop~\citep{bjorck2025gr00t} and RoboCasa-Kitchen~\citep{nasiriany2024robocasa}, we use the publicly available dataset\footnote{\scriptsize\url{https://huggingface.co/datasets/nvidia/PhysicalAI-Robotics-GR00T-X-Embodiment-Sim}}.
We use 1000 demonstrations per task for GR-1 tabletop, and 300 demonstrations per task for RoboCasa-Kitchen.
We provide a visualization of example tasks of our simulated benchmarks in~\Cref{fig:simtasks}.

\subsection{Training and evaluation details}
\label{appendix:simtrainingdetails}
\begin{figure}[h!]
\centering\small
\includegraphics[width=\linewidth]{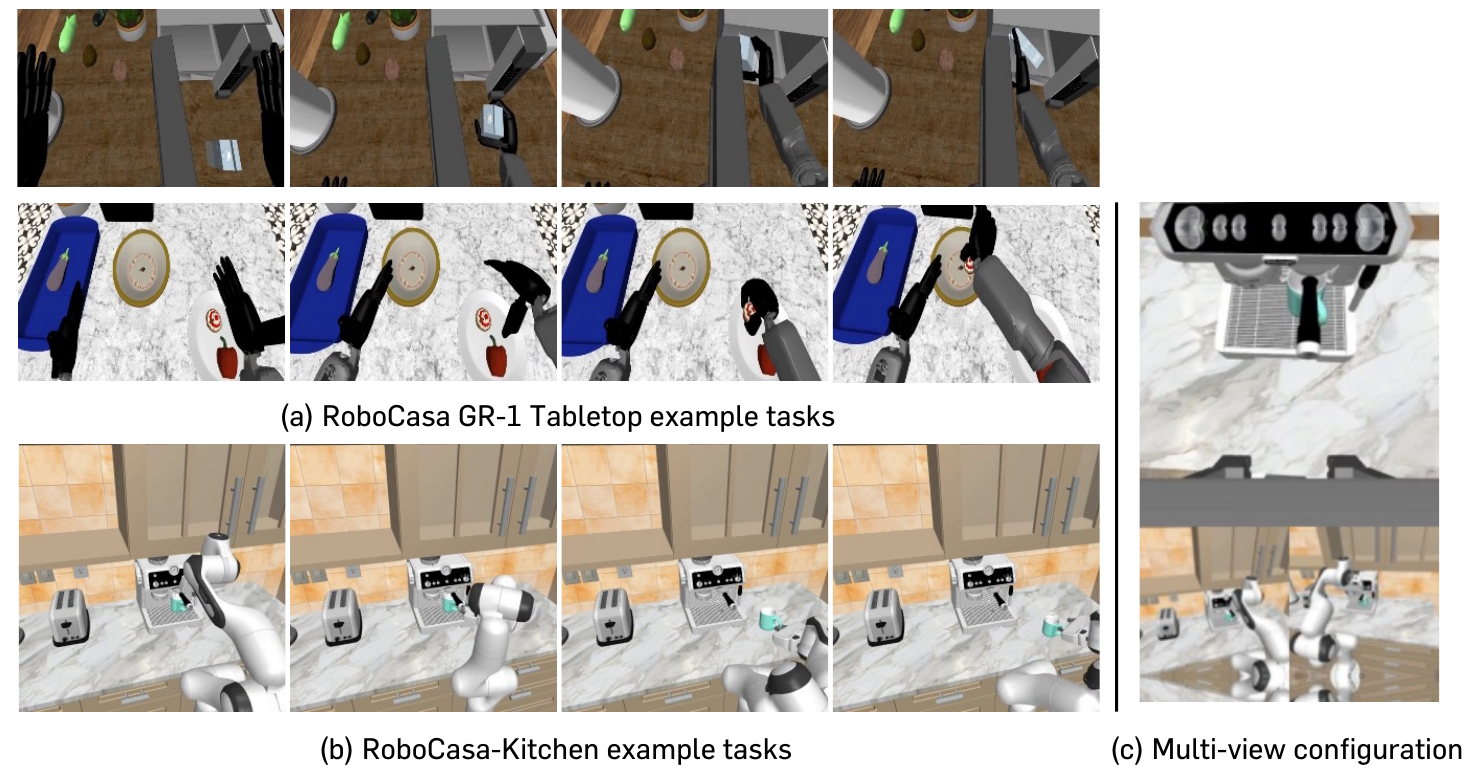}
\caption{Task examples of (a) RoboCasa GR-1 Tabletop, (b) RoboCasa-Kitchen, and (c) the multi-view observation input configuration.}
\label{fig:simtasks}
\end{figure}

For Cosmos3-Nano~\citep{agarwal2026cosmos} and Wan2.2-5B~\citep{wan2025wan}, we employ the publicly available checkpoints\footnote{\scriptsize\url{https://huggingface.co/nvidia/Cosmos3-Nano}}\footnote{\scriptsize\url{https://huggingface.co/Wan-AI/Wan2.2-TI2V-5B}}.
For Cosmos3 experiments, we train 16K optimization steps with a global batch size of 256 using AdamW with weight decay of 0.05, 500 steps of warmup to a peak learning rate of $2\times10^{-5}$, followed by linear decay to zero.
We train to generate 16 actions and 8 future video frames with a temporal stride of two, given a single observation frame.
For Fast-WAM~\citep{yuan2026fast} experiments, we train for 100K optimization steps with a global batch size of 64 using AdamW with weight decay of 0.01 and a peak learning rate of $10^{-4}$, with linear warmup over the first 5\% of training.
The input and output frames and the action horizon are identical to those of Cosmos3.
For RoboCasa-Kitchen, the three views (exterior left, exterior right, wrist) are stacked in a rectangular canvas, shown in~\Cref{fig:simtasks}.

For the evaluation protocol, we evaluate each method over 50 trials with fixed seeds for reproducibility. 
RoboCasa GR-1 Tabletop and RoboCasa-Kitchen both consist of 24 task suites, making 1200 evaluation trials for all models.
For Cosmos3, we use all 16 actions generated in a single forward process at execution without using classifier-free guidance, and we use 12 actions for Fast-WAM following the original work.

\subsection{Comparison Methods Details}

\vspace{0.01in}
\textbf{AGRA details.}~
AGRA~\citep{qiu2026making} applies REPA-style~\citep{yu2025representation} per-token cosine alignment to patch features from a frozen DINOv2 encoder~\citep{oquab2023dinov2}.
With the future latents fixed at the pure-noise level, we extract vision-token hidden states after the 8th transformer block and project them through a three-layer MLP with a hidden width of 2048.
For each temporal latent group, we use its first RGB frame as input to DINOv2 ViT-L/14 at $448\times448$ resolution, following the original work's implementations.
The resulting $32\times32$ patch-feature grid is bilinearly resized to the $8\times8$ latent grid for token-wise alignment.
We set the alignment loss weight to $0.5$.

\vspace{0.01in}
\textbf{Geometry Forcing details.}~
Geometry Forcing~\citep{wu2026geometry} aligns intermediate video-denoiser features with aggregator tokens from a frozen VGGT model~\citep{wang2025vggt}.
We adopt the angular alignment objective from the original work.
Specifically, we extract the value projections of vision tokens from the 8th transformer block and project them through a three-layer MLP with SiLU activations and a hidden width of 2048.
We minimize the mean per-token cosine distance ($1-\cos$) between the projected features and the output tokens of the final aggregator layer of VGGT-1B.
Following the original work, we set the alignment loss weight to $0.5$ and jointly train the projector and backbone.

\vspace{0.01in}
\textbf{VideoREPA details.}~
VideoREPA~\citep{zhang2026videorepa} aligns the pairwise cosine-similarity matrix of projected intermediate denoiser features with that of a frozen video foundation model using a margin-based hinge loss.
We extract hidden states after the 21st of 36 transformer blocks and project them through a three-layer MLP with a hidden width of 2048.
We retain the future video latent tokens and trilinearly resize the $2\times8\times8$ student feature grid to match the $4\times16\times16$ teacher tubelet grid.
The teacher receives clean future frames at $256\times256$ resolution.
We use VideoMAEv2 ViT-B~\citep{wang2023videomae} with a margin of $0.1$ and V-JEPA~2.1 ViT-L~\citep{mur2026v} with a margin of $0.05$.
Following the original work, we set the alignment loss weight to 0.5, half the flow-matching loss weight.

\vspace{0.01in}
\textbf{Tempered Self-similarity Alignment details.}~
Tempered Self-similarity Alignment (TSA)~\citep{kim2026tempered} converts pairwise cosine similarities into correspondence distributions using a temperature-scaled softmax over the spatial tokens of each target frame.
We use its masked variant (M-TSA), which restricts queries to the top $20\%$ of tokens in each temporal slice, ranked by patch-wise $\ell_1$ frame differences in the teacher input.
We reuse the VideoREPA pipeline with VideoMAEv2 ViT-B, replacing the alignment objective with the KL divergence from the teacher distribution to the student distribution, averaged over selected queries and target frames.
We set the temperature parameter to $\tau=0.1$ and set the alignment loss weight to 0.5.

\subsection{Analysis Details}
\label{appendix:analysisdetails}

\vspace{0.01in}
\textbf{Motion score analysis details.}~
We evaluate how well the latent motion score identifies each region in~\Cref{tab:motionscore} using the area under the receiver operating characteristic curve (AUROC). 
For each region, we treat locations inside the region as positives and those outside as negatives, using the motion score as the prediction score. Higher AUROC indicates that the score preferentially highlights the region, with 0.5 corresponding to chance-level discrimination.
The ground-truth evaluation regions are obtained from the segmentation masks provided by the simulator, where the moving arm, hand, and objects are selected based on centroid displacements of at least two pixels.

\vspace{0.01in}
\textbf{Attention ratio.}~
We define the attention ratio metric used in~\Cref{tab:policyanalysis} as the mean attention share assigned to a region divided by its mean spatial coverage:
\[
\mathrm{AttentionRatio}(R)
=
\frac{\mathbb{E}\!\left[\sum_{i=1}^{N} a_i c_{R,i}\right]}
{\mathbb{E}\!\left[\frac{1}{N}\sum_{i=1}^{N} c_{R,i}\right]},
\]
where \(a_i\) denotes action-token attention normalized over the \(N\) visual tokens within each temporal block, and \(c_{R,i}\in[0,1]\) is the fraction of token \(i\)'s spatial area covered by region \(R\). A value of one corresponds to uniform attention, while larger values indicate greater concentration relative to region size. 
We aggregate across evaluation windows and the conditioning and two future-latent blocks. 
We select the top four layers with the highest dynamic-region attention in the baseline model, and evaluate both models on the same layers to observe the dynamic region attention change.

\subsection{Hyperparameters and further details}
\label{appendix:implementationdetails}
For the motivation study in~\Cref{fig:motivatingstudy}, we set the temperature parameter $\beta$=2. 
For Motion-Guided Asymmetric Prediction (MoGAP), we set the auxiliary loss weight to  $\lambda_{\mathrm{MoGAP}}=1.0$.
For backbones using the Wan2.2 causal VAE tokenizer~\citep{wan2025wan}, the tokenizer first encodes a single-frame chunk and then encodes the remaining frames in four-frame chunks.
Because the first chunk has a different temporal extent, we compute the motion score for the first predicted latent using the forward difference
$m_{n,p}=\lVert\rvz_{t,n+1,p}-\rvz_{t,n,p}\rVert_2$.
To avoid division by zero when all motion scores vanish, we lower-bound the normalization denominator by $\delta=10^{-8}$. This yields $\widetilde m_{n,p}=0$ and uniform loss weights. We did not encounter this case in our experiments.
\newpage

\section{Real-Robot Experiments Details}
\label{appendix:realdetails}

\subsection{Hardware setup}
We use the OpenArm humanoid\footnote{\scriptsize\url{https://openarm.dev/}}, which consists of two 7-DoF arms, each equipped with a 6-DoF Inspire RH56F1 hand\footnote{\scriptsize\url{https://en.inspire-robots.com/product/rh56f1/}}. 
The robot head is mounted on a 2-DoF neck supporting pitch and yaw motions, with a ZED 2i stereo camera mounted on the neck.
During data collection, both the left and right egocentric camera views are recorded, while only the left view is used as visual input for training and evaluation. 
We use ExoArm-7, a wearable exoskeleton compatible with OpenArm, to teleoperate both robot arms, and Manus Pro gloves to track fingertip positions, from which finger joint angles are computed via inverse kinematics. 
The table height is kept fixed throughout data collection and evaluation.

\vspace{0.01in}
\subsection{Real-Robot Tasks}
We evaluate real-robot manipulation across six settings, comprising two in-domain settings and four generalization settings. The in-domain settings evaluate basic manipulation capabilities through pick-and-place and air-fryer manipulation. The generalization settings evaluate robustness to unseen objects, unseen environments, and dynamic object relocation under human interference.

\vspace{0.01in}
\textbf{In-domain tasks.}~
We consider two in-domain settings: pick-and-place and air-fryer manipulation. For pick-and-place, we evaluate three object--destination pairs: banana to basket, mango to air fryer, and apple to air fryer. 
The robot is instructed to pick up the target object and place it at the specified destination. 
Each pair is evaluated from three predefined initial object locations. For air-fryer manipulation, the robot is instructed to either open or close the air fryer, with the two operations evaluated separately (\Cref{fig:realworld_task_in_domain}).

\vspace{0.01in}
\textbf{OOD objects.}~
We evaluate object-level generalization by replacing the in-domain objects with four unseen objects which are not included in the training set: a box, a doll, a cube, and a tennis ball. Note that the doll used in this evaluation differs in shape and texture from the dolls seen during training.
The robot is instructed to pick up each object and place it into either the basket or the air fryer. 
We use the same three predefined initial locations as in the in-domain pick-and-place setting (\Cref{fig:realworld_task_ood_obj}).

\vspace{0.01in}
\textbf{OOD environments.}~
We evaluate environment-level generalization by varying the visual background while keeping the manipulation setup otherwise unchanged. Specifically, we change the tabletop covering to introduce backgrounds not observed in the in-domain setting (\Cref{fig:realworld_task_ood_env}).

\vspace{0.01in}
\textbf{OOD object relocation.}~
We evaluate robustness to dynamic object displacement by introducing human interference during the in-domain pick-and-place tasks. During execution, a human relocates the target object, requiring the robot to adapt to its updated position. We consider two difficulty levels. In set I, the object is relocated by a moderate distance at an earlier stage of the robot's approach. In set II, the object is displaced by a larger distance and the relocation occurs when the robot hand is closer to the object, requiring stronger adaptation to task-relevant visual changes (\Cref{fig:realworld_task_ood_obj_rel_compare}).

\subsection{Real-World Training and Evaluation Details}
\label{appendix:realtrainingdetails}

\begin{figure}[h!]
\centering\small
\includegraphics[width=\linewidth]{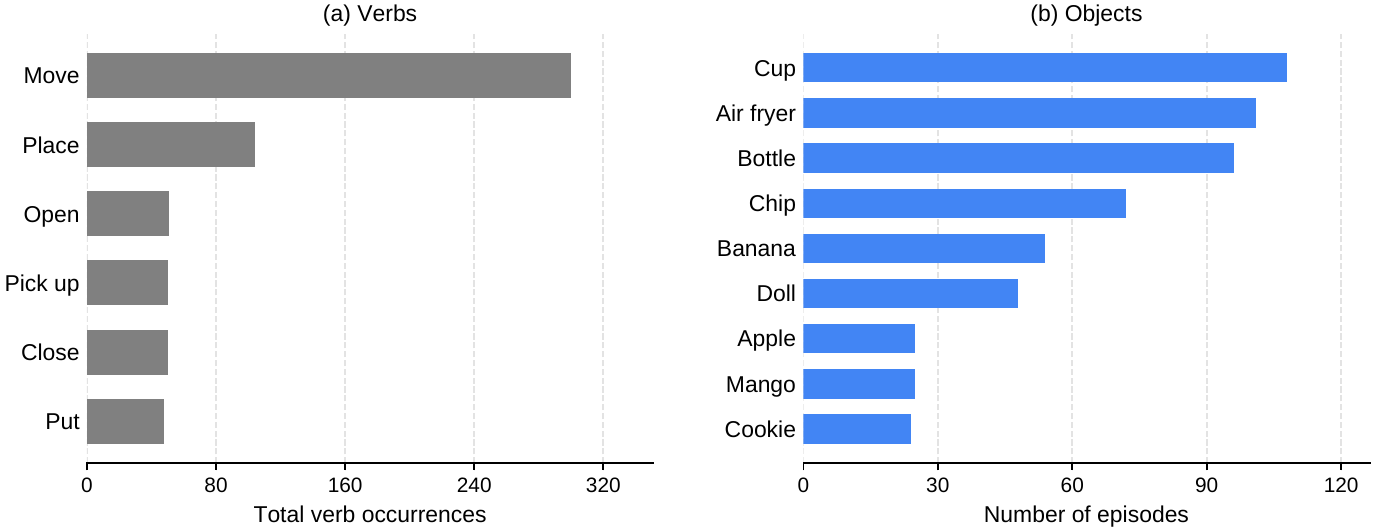}
\caption{Visualization of the OpenArm dataset distribution.}
\label{fig:dataset_distribution}
\end{figure}

\vspace{0.01in}
\textbf{Dataset.}~
We use a single dataset of teleoperated expert demonstrations across 21 tasks on our real-world setup, including the object pick-and-place and air fryer open-and-close in-domain tasks.
We visualize the verbs and objects of our OpenArm dataset in \Cref{fig:dataset_distribution}.

\vspace{0.01in}
\textbf{Training details.}~
For real-robot experiments, the dataset is substantially smaller than that of the simulated benchmark, and we find that conditioning on historical observations and applying image augmentations help mitigate overfitting. 
We adopt a video-to-video generation setup conditioned on four past frames sampled with a stride of two. 
For image augmentation, we randomly crop 5\% of the image and resize back to its original size. 
We also uniformly sample brightness, contrast, saturation factors, and hue shifts from \([0.7,1.3]\), \([0.6,1.4]\), \([0.5,1.5]\), and \([-0.08,0.08]\), respectively.
We train a single model for each method, for 8K optimization steps with a global batch size of 256.

\vspace{0.01in}
\textbf{Evaluation details.}~
We evaluate each policy over 24 trials per task, except for in-domain pick-and-place, where we use 48 trials to cover a more diverse set of objects, yielding 168 trials per method.
For pick-and-place, a trial is successful if the robot grasps the target object and places it at the designated destination. 
For air-fryer opening and closing, success requires moving the appliance to the fully open or closed state, respectively.

\input{figures/fig10_openarm_in_domain}
\input{figures/fig11_openarm_ood_obj} 
\input{figures/fig12_openarm_ood_env}
\input{figures/fig13_openarm_object_relocation_set_comparison}

\newpage

%% file: figures/fig10_openarm_in_domain.tex
\begin{figure}[h!]
\centering\small
\includegraphics[width=\linewidth]{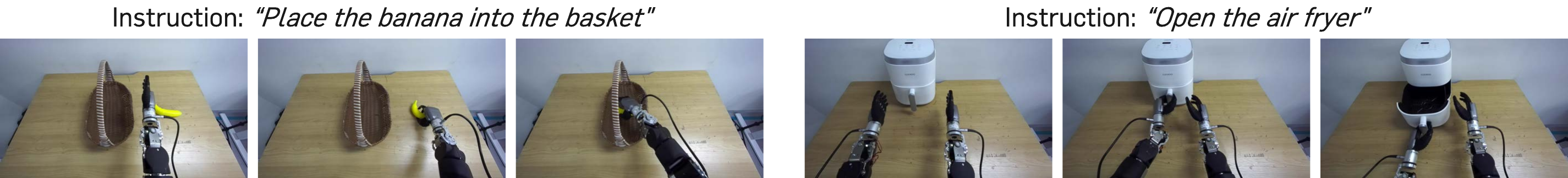}
\caption{In-domain task examples.}
\label{fig:realworld_task_in_domain}
\end{figure}

%% file: figures/fig11_openarm_ood_obj.tex
\begin{figure}[h!]
\centering\small
\includegraphics[width=\linewidth]{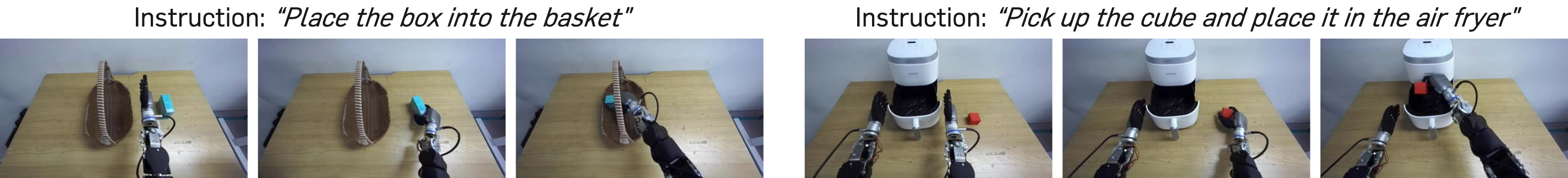}
\caption{Unseen objects task examples.}
\label{fig:realworld_task_ood_obj}
\end{figure}

%% file: figures/fig12_openarm_ood_env.tex
\begin{figure}[h!]
\centering\small
\includegraphics[width=\linewidth]{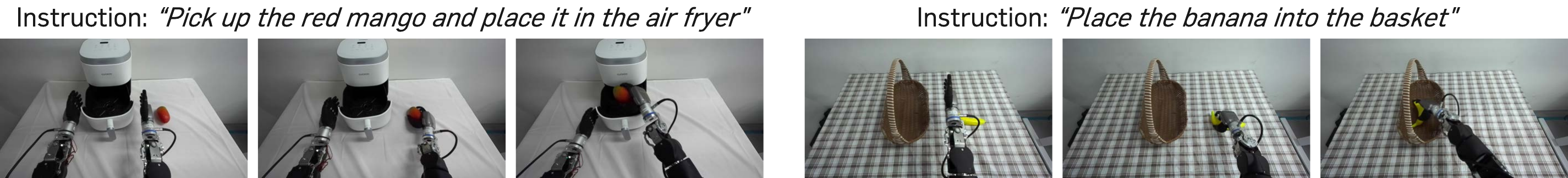}
\caption{Unseen environments task examples.}
\label{fig:realworld_task_ood_env}
\end{figure}

%% file: figures/fig13_openarm_object_relocation_set_comparison.tex
\begin{figure}[h!]
\centering\small
\includegraphics[width=\linewidth]{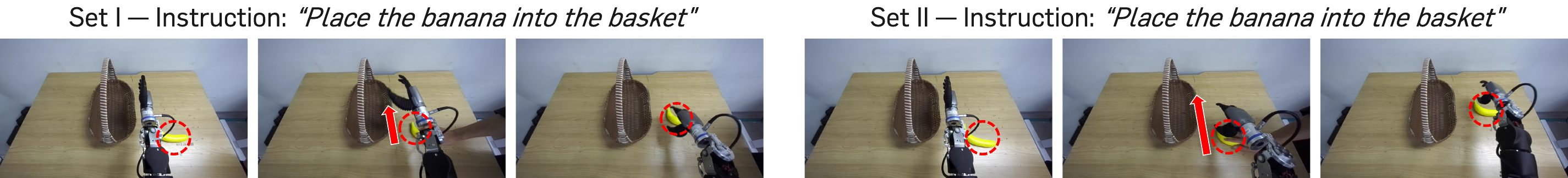}
\caption{Object relocation task examples (left: Set I; right: Set II)}
\label{fig:realworld_task_ood_obj_rel_compare}
\end{figure}